\documentclass[nonatbib]{article}

\usepackage[final]{neurips_2024}

\usepackage[utf8]{inputenc} 
\usepackage[T1]{fontenc}    
\usepackage{hyperref}       
\usepackage{url}            
\usepackage{booktabs}       
\usepackage{amsfonts}       
\usepackage{nicefrac}       
\usepackage{microtype}      
\usepackage{xcolor}         
\usepackage{graphicx} 
\usepackage{subcaption}
\usepackage{tabularx}
\usepackage{booktabs}
\usepackage{cite}
\usepackage{amsmath}
\usepackage[numbers]{natbib}
\usepackage{multirow}
\usepackage{amssymb}
\title{LiteVAR: Compressing Visual Autoregressive Modelling with Efficient Attention and Quantization}

\author{%
  Rui Xie$^{1,2}$, Tianchen Zhao$^{1,2}$, Zhihang Yuan$^{1,2}$, Rui Wan$^{3}$, Wenxi Gao$^{1}$, 
  \And
  Zhenhua Zhu$^{1}$, Xuefei Ning$^{1,2}$, Yu Wang$^{1,2}$
}

\begin{document}

\maketitle

\vspace{-20pt}

\begin{center}
{\fontsize{8.5pt}{15pt}\selectfont
$^1$Tsinghua University~~~~$^2$Infinigence AI~~~~$^3$Fudan University
}
\vspace{10pt}
\end{center}

\begin{abstract}

Visual Autoregressive (VAR) has emerged as a promising approach in image generation, offering competitive potential and performance comparable to diffusion-based models. However, current AR-based visual generation models require substantial computational resources, limiting their applicability on resource-constrained devices. To address this issue, we conducted analysis and identified significant redundancy in three dimensions of the VAR model: (1) the attention map, (2) the attention outputs when using classifier free guidance, and (3) the data precision. Correspondingly, we proposed efficient attention mechanism and low-bit quantization method to enhance the efficiency of VAR models while maintaining performance. 
With negligible performance lost (less than 0.056 FID increase), we could achieve 85.2\% reduction in attention computation, 50\% reduction in overall memory and 1.5x latency reduction. 
To ensure deployment feasibility, we developed efficient training-free compression techniques and analyze the deployment feasibility and efficiency gain of each technique. 

\end{abstract}

\section{Introduction}

Visual Autoregressive (VAR~\cite{tian2024visual}) modeling has explored the autoregressive (AR) paradigm for visual generation, achieving performance comparable to state-of-the-art diffusion models. By leveraging the "multi-scale" nature of images, VAR introduces a scale-by-scale generation scheme, progressing from coarse to fine. However, despite operating on high-level visual tokens, the VAR generation process still requires iterative token generation across multiple scales, resulting in substantial computational cost. This challenge hinders the broader application of VAR models on resource-constrained platforms, highlighting the need for efficiency improvements. In this paper, we focus on designing training-free model compression techniques to reduce the computational and memory burden of VAR models. We hope our research could shed some lights on practical acceleration of VAR and even more AR-based image generative models~\cite{llamagen,luminaX}.

\begin{figure}
    \centering
    \includegraphics[width=0.95\linewidth]{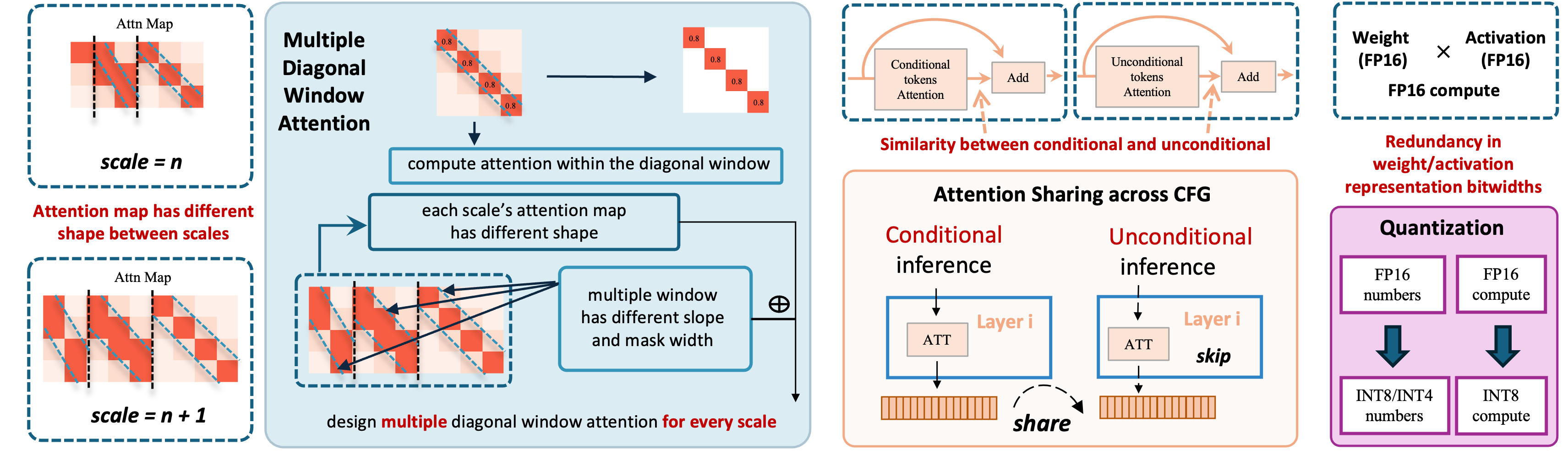}
    \caption{\textbf{Three dimensions of redundancy and corresponding compression techniques.} 
    We discover redundancy exists in the attention map level, the classifier free guidance level, and the representation data precision level. We design the multi-diagonal windowed attention, CFG-wise sharing, and mixed precision quantization to address the above redundancy.  
    }
    \label{fig:framework}
    \vspace{-21pt}
    
\end{figure}

Based on algorithmic characteristics, we explore the redundancy for VAR to design corresponding optimization. As presented in Fig.~\ref{fig:framework}, we conclude the redundancy in the following dimensions: 

\textbf{Redundancy In Attention Map.} As discussed in prior literature on vision transformer~\cite{swin}, visual models tend to exhibit a local feature extraction nature. Using global attention that aggregates all tokens may therefore be redundant, with much of the computation spent on representing relatively weak long-range relationships between visual tokens. Inspired by this, we visualize the attention map of the VAR model in Fig.~\ref{fig:combine}-(a) and find that tokens primarily focus on their local window in the attention map, while most attention values for distant tokens are close to zero. Additionally, we observe a unique ``multi-diagonal'' pattern in the VAR attention map, where visual tokens are locally aggregated within each scale.

In order to leverage the unique characteristics of VAR attention maps,  we propose replacing global attention with windowed local attention at each stage, which we term ``multidiagonal windowed attention''. This approach effectively reduces both the computational and memory costs of attention. By incorporating multi-diagonal windowed attention, we could save 70-80\% of attention computation without compromising performance. 
While attention computation is not a critical bottleneck in the current experimental setting (VAR on ImageNet 256x256) due to the relatively low resolution, it is important to note that attention costs scale quadratically with token length. A recent research~\cite{yuan2024ditfastattn} suggests that for 2K resolution generation, attention computation can become the primary bottleneck.

\textbf{Redundancy In Attention Outputs When Using Classifier-Free Guidance (CFG).} The CFG technique~\cite{ho2022classifier} is widely applied in conditioned generation, not only for diffusion models but also for autoregressive (AR) models~\cite{tian2024visual,mentzer2023fsq}. In this technique, the model is run twice—once with and once without the control signal—and the outputs are combined via a weighted sum. The weighting coefficient controls the strength of the control signal. Recent studies~\cite{yuan2024ditfastattn} have identified computational redundancy between the conditional and unconditional inferences in diffusion models. In this work, we investigate whether similar redundancy exists in AR models, using VAR as a representative example.
By visualizing the similarity between the attention QKV of the conditional and unconditional branches in VAR generation (in Fig.~\ref{fig:combine})-(b), we observed significant overlap across different blocks, heads, and scales.
For leveraging this redundancy, following previous work, we propose sharing the attention output between the conditional and unconditional branches, thereby skipping the computation for one branch.
Combining multidiagonal windowed attention with the CFG sharing technique, we could reduce 85-90\% of attention computation. 

\textbf{Redundancy In Data Precision.} Prior low-bit quantization methods~\cite{jacob2018quantization,nagel2021white} reveal that the high precision floating-point (FP) representation for neural network weight and activation are redundant. The Post Training Quantization (PTQ) has proven to be an effective method for both reducing model size, memory footprint, and computational complexity. Following recent advances in diffusion visual generation model quantization~\cite{yuan2022ptq4vit, zhao2024vidit}, we apply post training quantization technique to VAR models. Although W8A8QKV8 quantization achieves satisfying performance. We empirically witness notable visual quality degradation for lower bit-width (W6A6 and W4A8). Furthermore, we discover that the quantization is ``bottlenecked'' by some highly sensitive layers under lower bit-width, and adopt mixed precision quantization method to preserve these highly sensitive layers at higher bit-width. 

We summarizes the knowledge of our redundancy analysis, the performance-efficiency trade-off, and deployment feasibility of existing methods in Sec~\ref{sec:5.2}.

\section{Attention Redundancy: Multi-Diagonal Window Attention (MDWA) }

\begin{figure}
    \centering
    \begin{subfigure}[b]{0.6\textwidth}
        \includegraphics[width=\textwidth]{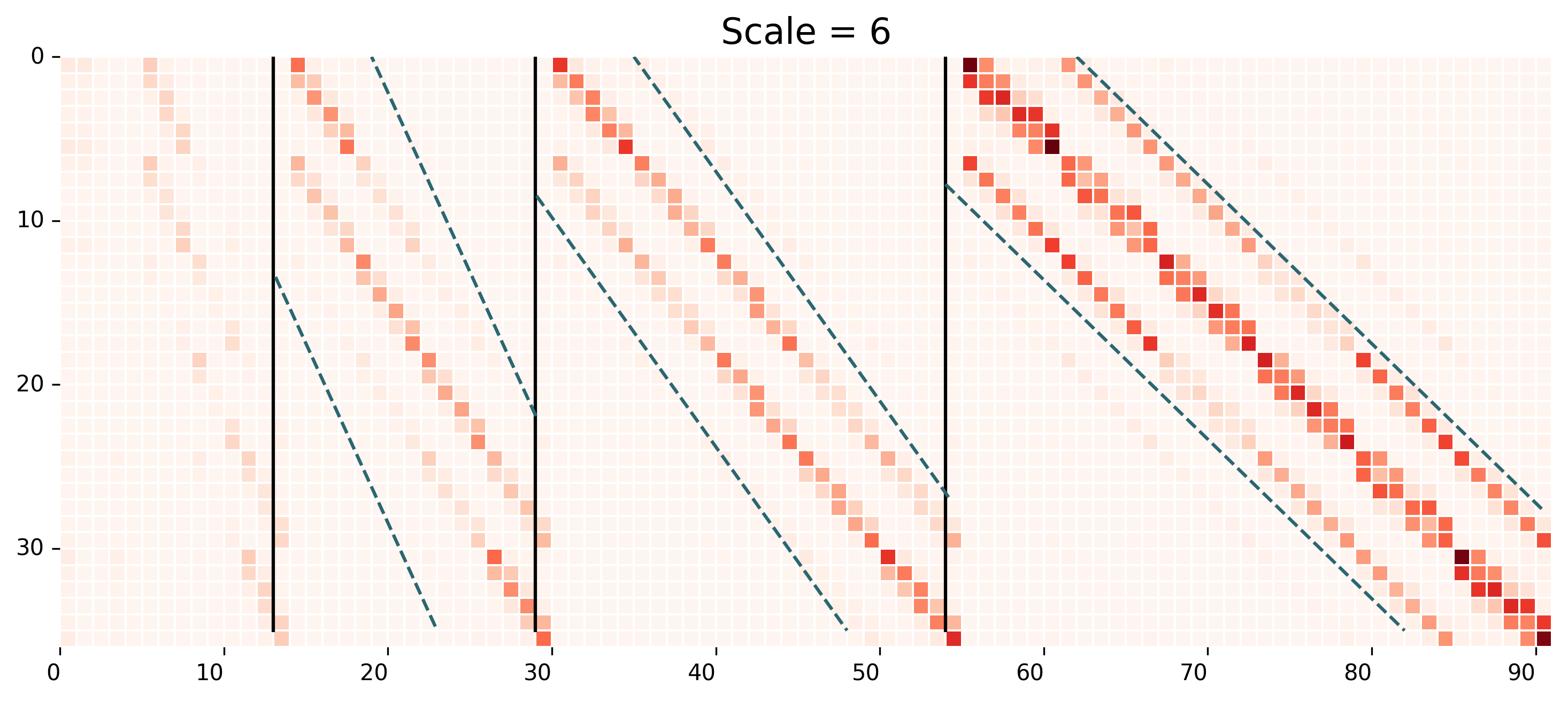}
        \caption{}
    \label{fig:attn_vis}
    \end{subfigure}%
    \hspace{0pt} 
    \begin{subfigure}[b]{0.35\textwidth}
        \includegraphics[width=\textwidth]{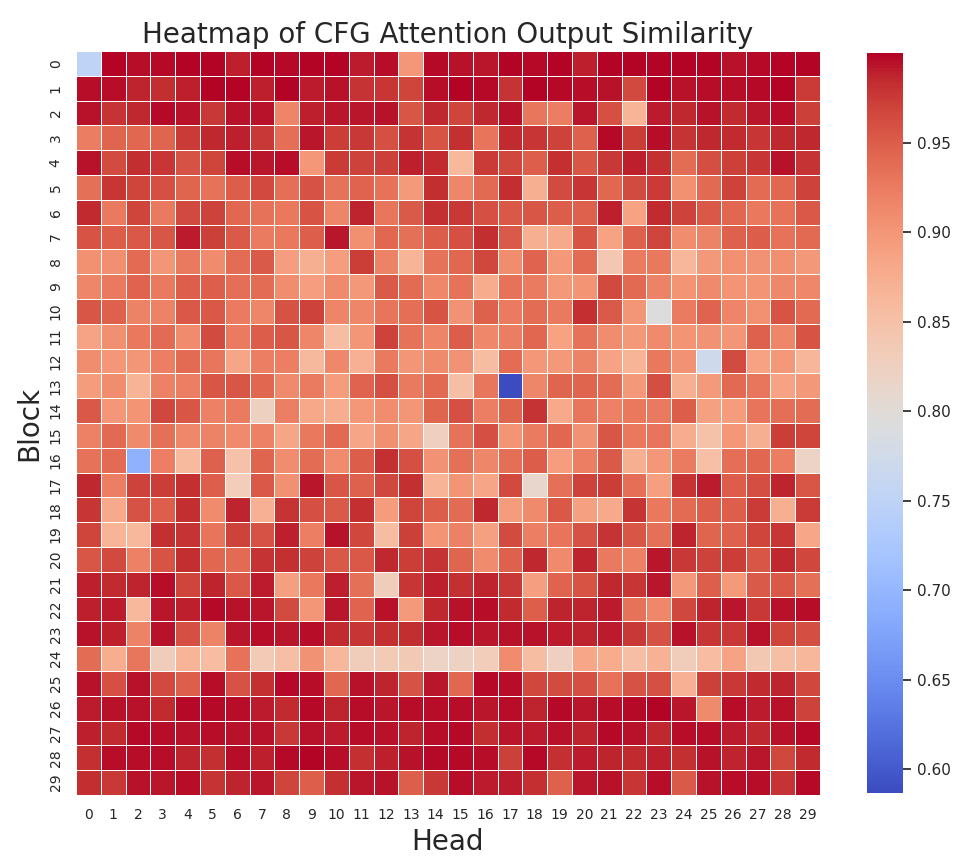}
        \caption{}
    \label{fig:cfg_similar}
    \end{subfigure}%
    \caption{\textbf{Attention map characteristics. }(a) \textbf{Multi-diagonal concentration.} VAR model's attention values are concentrated on multiple diagonals, with each diagonal exhibiting a distinct shape across different scales. Consequently, we have designed a separate window attention mechanism for each scale, which we refer to as Multi-Diagonal Window Attention (MDWA). (b) \textbf{Similarity} of Attention Outputs between \textbf{Conditional} and \textbf{Unconditional} Generation.}
    \label{fig:combine}
\end{figure}


We visualize the attention map for VAR models in Fig.~\ref{fig:combine}-(a). As shown, for most tokens, the attention map concentrates within local regions at each scale, with more than 80\% of the values representing interactions between spatially distant visual tokens being close to zero. Therefore, replacing the original global attention with local windowed attention can significantly reduce computation while preserving the majority of meaningful values in the attention map. Leveraging the unique multi-diagonal characteristics of the VAR attention map, we propose a specialized multi-diagonal windowed attention (MDWA) pattern to compress redundancy at the attention map level.

Specifically, considering the VAR model with $K$ scales, each scale containing $s_k^2$ tokens. For the k-th scale, the attention mechanism aggregates tokens from the current scale ($s_k^2$) with all tokens from previous scale ($\sum_1^k s_i^2$). For example, when $k=2$, the attention map $X$ has a shape of $[4,5]$, where 4 represents the number of visual tokens at the current scale ($2^2$), and 5 represents the total tokens from previous scales ($1^2+2^2$). As shown in Fig.\ref{fig:combine}-(a), we separate the attention map into $N$ parts (indicated by vertical black lines) and design a local windowed attention pattern (marked by blue lines) with window width $w$. We introduce a metric, $R_w$, to control the trade-off between performance and efficiency. The $R_w$ is defined as the division of the summation of all elements within the window, with respect to the summation of all values in the current part. Since the attention values are within range $[0,1]$, the value of $R_w$ could be interpreted as the measurement of ``how many percentage of dominant attention values are contained in the local window''. We gradually increase the window size from zero until $R_w$ reaches a specific pre-defined ratio $R_0$ (e.g., 0.95).  Table~\ref{tab:1} presents the performance efficiency trade-off with different $R_0$. When $R_0=1$, the attention pattern falls back to full attention. We further provide the detailed process of the MDWA pattern design.

(1) We perform model inference on a subset of the training data and save the attention maps (after softmax) as a reference for designing the attention pattern.

(2) Given an attention map at the k-th scale with the shape $[s_k^2, \sum_1^k s_i^2]$, we first divide it into $k-2$ parts, where the first part contains $[s_k^2, \sum_1^3 s_i^2]$, and the rest j-th part has the shape $[s_k^2, s_{k-j+1}^2]$. For each part, we gradually increase the window size $w$ until the ratio $R_w$ reaches a predefined value $R_0$.

(3) This process is repeated to determine the optimal window size for each scale, block, and head of the attention map.

\label{sec:exp_mdwa}


\textbf{Image Evaluation Settings.} 
We adopt FID~\cite{heusel2017gans}, IS~\cite{salimans2016improved} for fidelity evaluation, and ImageReward~\cite{xu2024imagereward} for human preference. Following the original VAR code implementation, we use the 10-scale VAR with a CFG scale of 4. We generate 8K images on the ImageNet dataset to ensure the stability of the metric scores. 

\textbf{MDWA implementation details.} In the original VAR design, the $s_k$ values for the 10 scales are $(1, 2, 3, 4, 5, 6, 8, 10, 13, 16)$. We collect 80 samples in the training set and save their attention maps as reference. The multi-diagonal windowed attention patterns are designed following the aforementioned process. Additionaly, through analyzing the distribution of attention values, we observe that in the initial parts of the attention map, certain tokens occasionally exhibit uniformly high attention values across all tokens. This aligns with the "attention sink" phenomenon described in prior literature~\cite{attention_sink}. Since the computational cost of these initial parts is relatively low, we retain the full attention pattern for the first three parts of the attention map.

\textbf{Experimental Results.}
The width of our designed multi-diagonal window attention mechanism was determined by a threshold setting. We tested different threshold values, including 0.95, 0.9, 0.85, 0.8, 0.7, and 0.6, and evaluated the image quality generated under each threshold. We generated 8k ImageNet images for evaluation, as shown in Table \ref{tab:1}. The threshold of 0.95 yielded the best results, while a threshold of 0.6 still produced acceptable image quality. 

\begin{table}[]
\centering
\caption{\textbf{Performance of MDWA for different Threshold on ImageNet.} Image quality evaluation and Calculation saving for different \textbf{Threshold} settings in Multi-Diagonal Window Attention.}
\begin{tabular}{@{}ccccc@{}}
\toprule[1pt]
\textbf{Threshold} & \textbf{FLOPs Saving(\%)} & \textbf{FID($\downarrow$)} & \textbf{IS($\uparrow$)} & \textbf{Image Reward($\uparrow$)} \\ \midrule
1                  & 0.00                            & 13.39                      & 257.34                  & -0.28                             \\
0.95               & 70.34                           & 13.47                      & 260.95                  & -0.28                             \\
0.90               & 73.43                           & 13.50                      & 261.45                  & -0.29                             \\
0.85               & 75.47                           & 13.72                      & 259.54                  & -0.31                             \\
0.80               & 76.82                           & 13.77                      & 258.45                  & -0.34                             \\
0.70               & 79.36                           & 13.94                      & 254.17                  & -0.40                             \\
0.60               & 81.39                           & 14.39                      & 250.97                  & -0.48                             \\ \bottomrule[1pt]
\end{tabular}
\label{tab:1}
\end{table}

\section{CFG Redundancy: Attention Sharing across CFG (ASC)}


Classifier-free guidance (CFG) is widely used for conditional generation~\cite{saharia2022photorealistic}\cite{ramesh2022hierarchical}\cite{ho2022classifier}, requiring two model inferences: one with the condition signal and one without. Previous research~\cite{yuan2024ditfastattn} has explored reducing the redundancy from the similarity between conditional and unconditional inferences in diffusion models. Building on this, we investigate similar redundancy in AR-based image generation. As shown in Fig.\ref{fig:combine}-(b), we observe high similarity between the attention maps of conditional and unconditional inferences. Based on this, we propose the Attention Sharing across CFG (ASC) technique, which reuses the attention output from the conditional inference for the unconditional inference, significantly reducing attention computation cost. Since the vast majority of layers exhibit high attention map similarity, we reuse the attention maps across the entire network. We will further explore selectively reusing maps in layers with higher similarity to balance performance and efficiency in future work. 






\textbf{Experimental Results.} We applied the Attention Sharing across CFG (ASC) technique with the MDWA technique, the results, as presented in Table \ref{tab:ablation}. The generated images indicate that the loss introduced by ASC is minimal. In fact, for some metrics, ASC even outperformed the non-shared attention computation, demonstrating its effectiveness. Combining the MDWA with ASC, we could achieve 85\%-90\% attention computation savings with negligible visual quality degradation. 

\begin{figure}
    \centering
    \begin{subfigure}[b]{0.17\textwidth}
        \includegraphics[width=\textwidth]{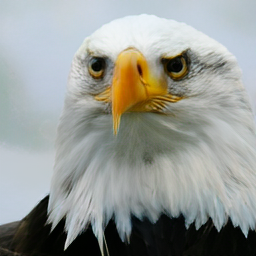}
        \caption{original}
    \end{subfigure}%
    \hspace{0pt} 
    \begin{subfigure}[b]{0.17\textwidth}
        \includegraphics[width=\textwidth]{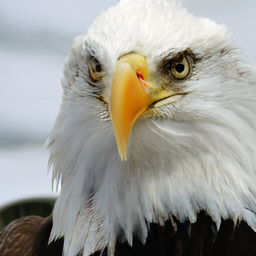}
        \caption{MDWA}
    \end{subfigure}%
    \hspace{0pt} 
    \begin{subfigure}[b]{0.17\textwidth}
        \includegraphics[width=\textwidth]{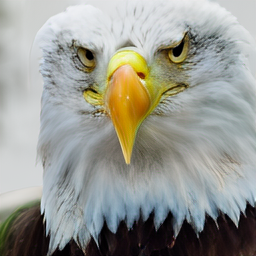}
        \caption{MDWA+ASC}
    \end{subfigure}

    \caption{Comparison of original image generation with the techniques of Multi-Diagonal Window Attention(MDWA) and CFG-wise attention sharing(ASC).}
    \label{fig:mask}
    \vspace{-10pt}
\end{figure}


\section{Data Precision Redundancy: Mixed
Precision Quantization}


Post Training Quantization(PTQ) has proven to be an efficient and effective model compression method ~\cite{nagel2021white}. It converts the floating-point data into low-bit integers,the process could be represented as:
$$x_q = \text{round}(\text{clamp}((x-z)/s, -2^{B-1}, 2^{B-1}))$$
The $s$ (scale) and $z$ (zero point) are quantization parameters, which are determined offline based on stored calibration data with:
$$s = \text{max}(\text{abs}(x))$$
$$z = (\text{max}+\text{min})/2$$
However, we empirically observe that using this straightforward quantization method leads to significant quality degradation, even at W8A8QKV8 (weights, activation, and the QKV in attention are quantized to 8-bit integers). Building on recent advancements in language model quantization~\cite{xiao2023smoothquant}\cite{yao2022zeroquant}, we adopt dynamic quantization parameters for activation quantization, where $s$ and $z$ are computed online to adapt to diverse activations. Since calculating these quantization parameters only requires obtaining the maximum and minimum values of the data, the additional computational cost remains minimal. We apply this dynamic quantization scheme to VAR models, with results presented in Table~\ref{tab:mixprecision}. 

While achieving W8A8QKV8 quantization without performance loss, we still observe quality degradation at lower bit-widths (e.g., W4A8QKV8). To investigate the cause, we analyzed the model and found that quantizing certain layers leads to significant performance drops, while others do not. This reveals that highly quantization-sensitive layers create a bottleneck for low-bit quantization. As shown in Fig.~\ref{fig:ffn.fc2} in the appendix, our extensive analysis of the VAR model layers indicates that quantizing the "ffn.fc2" layer to W4A6 causes a disproportionately larger quality degradation compared to other layers. To address this "bottleneck phenomenon", we propose employing mixed precision quantization, maintaining higher bit-widths for these particularly sensitive layers.

\begin{figure}[ht]
    \centering
    \begin{subfigure}[b]{0.18\textwidth}
        \includegraphics[width=\textwidth]{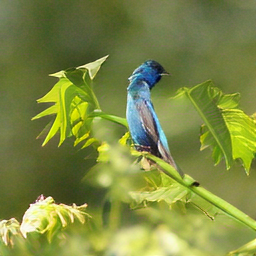}
        \caption{fp16}
    \end{subfigure}%
    \hspace{0pt} 
    \begin{subfigure}[b]{0.18\textwidth}
        \includegraphics[width=\textwidth]{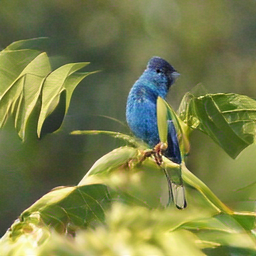}
        \caption{w8a8}
    \end{subfigure}%
    \hspace{0pt} 
    \begin{subfigure}[b]{0.18\textwidth}
        \includegraphics[width=\textwidth]{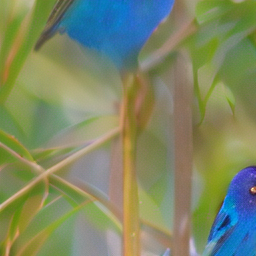}
        \caption{w4a8}
    \end{subfigure}%
    \hspace{0pt} 
    \begin{subfigure}[b]{0.18\textwidth}
        \includegraphics[width=\textwidth]{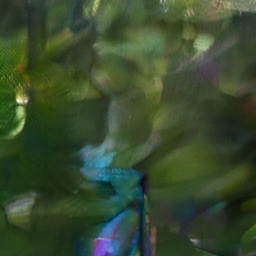}
        \caption{w6a6}
        
    \end{subfigure}

    \vspace{0.2cm} 
    
    \begin{subfigure}[b]{0.18\textwidth}
        \includegraphics[width=\textwidth]{figure/fp16_o_1.png}
        \caption{fp16}
    \end{subfigure}%
    \hspace{0pt} 
    \begin{subfigure}[b]{0.18\textwidth}
        \includegraphics[width=\textwidth]{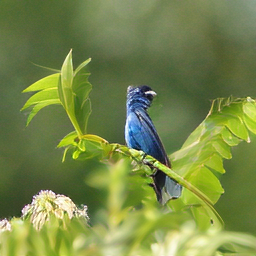}
        \caption{w8a8+MP}
    \end{subfigure}%
    \hspace{0pt} 
    \begin{subfigure}[b]{0.18\textwidth}
        \includegraphics[width=\textwidth]{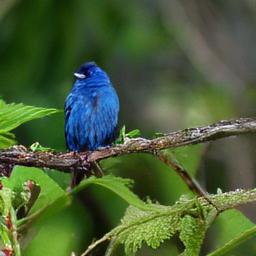}
        \caption{w4a8+MP}
    \end{subfigure}%
    \hspace{0pt} 
    \begin{subfigure}[b]{0.18\textwidth}
        \includegraphics[width=\textwidth]{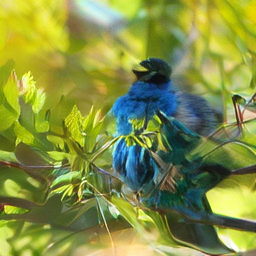}
        \caption{w6a6+MP}
        
    \end{subfigure}
    \caption{\textbf{Comparison of original image, quantized image and quantized image with protection of sensitive layers.} Top row: Naive quantized image exhibit substantial blurring or loss of legible content. Bottom row: A significant improvement in image quality post-quantization.}
    \label{fig:quantization}
\end{figure}


\textbf{Quantization Scheme.}
We adopt the simple min-max quantization scheme. The quantization parameters for activation are dynamic and computed online with negligible overhead. The mixed precision plan are determined offline based on the calibration data.


\textbf{Experimental Results.} The evaluation scheme are kept consistent with Sec.\ref{sec:exp_mdwa}. 
As shown in Table \ref{tab:mixprecision} and Fig.~\ref{fig:quantization}, both W8A8 and W8A8QKV8 exhibit no performance loss, generating images nearly identical to those produced with FP16. However, the images generated by W4A8 and W6A6 show noticeable blurring, underscoring the need for mixed precision quantization. By adopting mixed precision quantization, both W4A8 and W6A6 experience significant improvements in visual quality and metric scores. In fact, W4A8 with mixed precision can achieve nearly the same generation quality as uniform W8A8 quantization.

\begin{table}[]
\centering
\caption{Performance of image generation on ImageNet under various settings of quantization. Mixed-precision design significantly improves the performance under low bitwidth quantization.}

\begin{tabular}{@{}ccccc@{}}
\toprule[1pt]
\textbf{\begin{tabular}[c]{@{}c@{}}Bit-width\\ (W/A/QKV)\end{tabular}} & \textbf{Mix-Precision (FP16)} & \textbf{FID($\downarrow$)} & \textbf{IS($\uparrow$)} & \textbf{Image reward($\uparrow$)} \\ \midrule \midrule
16/16/16                                                               & $-$                          & 13.39                      & 257.34                  & -0.28                             \\ \midrule
\multirow{2}{*}{8/8/8}                                                 & $-$                          & \textbf{12.71}             & \textbf{249.04}         & \textbf{-0.33}                    \\
                                                                       & $\checkmark$                 & 13.08                      & 253.43                  & -0.30                             \\ \midrule
\multirow{2}{*}{4/8/8}                                                 & $-$                          & 54.29                      & 40.71                   & -1.43                             \\
                                                                       & $\checkmark$                 & \textbf{12.82}             & \textbf{228.59}         & \textbf{-0.41}                    \\ \midrule
\multirow{2}{*}{6/6/8}                                                 & $-$                          & 66.53                      & 26.08                   & -1.68                             \\
                                                                       & $\checkmark$                 & 18.54                      & 133.13                  & -0.75                             \\ \midrule
4/6/8                                                                  & $-$                          & 111.24                     & 9.79                    & -2.10                             \\ \midrule
4/4/8                                                                  & $-$                          & 133.38                     & 6.63                    & -2.15                             \\ \bottomrule[1pt]
\end{tabular}

\label{tab:mixprecision}
\end{table}

\section{Analysis}

\subsection{Ablation Studies}

As demonstrated in Fig.~\ref{fig:mask}, the introduction of MDWA and ASC results in only a slight performance degradation (+0.05 FID). Furthermore, replacing the uniform W4A8QKV8 quantization with a mixed precision scheme significantly reduces performance loss. LiteVAR maintains performance comparable to the FP16 baseline while effectively compressing redundancy across three dimensions.

\subsection{Takeaways for VAR Compression Techniques}
\label{sec:5.2}

\textbf{Efficiency Improvement.} The MDWA and CFG-sharing could reduce 85\%-90\% attention computation and reduce 80\% attention map activation memory cost with negligible computational cost. Although for current application (ImageNet 256$\times$256), the attention computation and attention map memory cost is not excessive. However, the attention computation and memory cost grows quadratically with the token length. For higher resolution (2K) generation, the attention operation becomes the major bottleneck. In such case, the efficient attention mechanism could significantly reduce the computation cost (69.6\% of the FLOPs), and the memory cost for saving the attention map (31.07GB). The quantization could effectively reduce both the computational cost and memory cost of the model. Taking W8A8 as an example, it could reduce 2$\times$ of model memory, and achieve around 1.5$\times$ latency speedup. 


\textbf{Efficiency of Compression Methods.} In addition to the efficiency improvement that the compression method brings, the efficiency of the compression method itself is also critical for practical application. Therefore, we design \textbf{training-free} compression techniques. Unlike many pruning-based methods that require model fine-tuning, MDWA attention compression eliminates the need for additional training or large-scale data. Similarly, for post-training quantization, we employ an efficient scheme that does not rely on gradient-based optimization of quantization parameters.

\textbf{Deployment Feasibility.}
The CFG-sharing technique requires no additional hardware support to implement, while the MDWA and quantization requires customized CUDA kernels to achieve speedup and memory savings. For the low-bit quantization, we adopt the commonly used minmax dynamic quantization scheme, which is supported by many deployment frameworks~\cite{atom, qserve}. The mixed precision quantization also does not requires additional support other than the W4A8 kernel (which is also supported by mainstream deployment frameworks).

\begin{table}[]
\centering
\caption{\textbf{Ablation studies of LiteVAR techniques.} When gradually incorporating LiteVAR's techniques, compressing attention by 85\% and reducing the bit width to W4A8QKV8, the generated images are acceptable.}
\begin{tabular}{@{}cccccc@{}}
\toprule[1pt]
\multicolumn{3}{c}{\textbf{Method}}                   & FID            & IS           & ImageReward  \\ \cmidrule(r){1-3}
MDWA         & ASC          & Quant(W/A/QKV) & ($\downarrow$) & ($\uparrow$) & ($\uparrow$) \\ \midrule \midrule
$-$     & $-$     & 16/16/16       & 13.39          & 257.34       & -0.28        \\ \midrule
$\checkmark$ & $-$     & 16/16/16       & 13.47          & 260.95       & -0.28        \\
$\checkmark$ & $\checkmark$ & 16/16/16       & 13.45          & 248.8        & -0.27        \\
$\checkmark$ & $\checkmark$ & 4/8/8          & 52.27          & 33.87        & -1.6         \\
$\checkmark$ & $\checkmark$ & 4/8/8+MP       & 13.34          & 224.74       & -0.39        \\ \bottomrule[1pt]
\end{tabular}
\label{tab:ablation}
\end{table}

\clearpage


\bibliographystyle{plain} 
\bibliography{neurips_2024} 

\begin{thebibliography}{10}

\bibitem{heusel2017gans}
Martin Heusel, Hubert Ramsauer, Thomas Unterthiner, Bernhard Nessler, and Sepp Hochreiter.
\newblock Gans trained by a two time-scale update rule converge to a local nash equilibrium.
\newblock {\em Advances in neural information processing systems}, 30, 2017.

\bibitem{ho2022classifier}
Jonathan Ho and Tim Salimans.
\newblock Classifier-free diffusion guidance.
\newblock {\em arXiv preprint arXiv:2207.12598}, 2022.

\bibitem{jacob2018quantization}
Benoit Jacob, Skirmantas Kligys, Bo~Chen, Menglong Zhu, Matthew Tang, Andrew Howard, Hartwig Adam, and Dmitry Kalenichenko.
\newblock Quantization and training of neural networks for efficient integer-arithmetic-only inference.
\newblock In {\em Proceedings of the IEEE conference on computer vision and pattern recognition}, pages 2704--2713, 2018.

\bibitem{qserve}
Yujun Lin, Haotian Tang, Shang Yang, Zhekai Zhang, Guangxuan Xiao, Chuang Gan, and Song Han.
\newblock Qserve: W4a8kv4 quantization and system co-design for efficient llm serving.
\newblock {\em arXiv preprint arXiv:2405.04532}, 2024.

\bibitem{swin}
Ze~Liu, Yutong Lin, Yue Cao, Han Hu, Yixuan Wei, Zheng Zhang, Stephen Lin, and Baining Guo.
\newblock Swin transformer: Hierarchical vision transformer using shifted windows.
\newblock In {\em Proceedings of the IEEE/CVF international conference on computer vision}, pages 10012--10022, 2021.

\bibitem{mentzer2023fsq}
Fabian Mentzer, David Minnen, Eirikur Agustsson, and Michael Tschannen.
\newblock Finite scalar quantization: Vq-vae made simple.
\newblock {\em arXiv preprint arXiv:2309.15505}, 2023.

\bibitem{nagel2021white}
Markus Nagel, Marios Fournarakis, Rana~Ali Amjad, Yelysei Bondarenko, Mart Van~Baalen, and Tijmen Blankevoort.
\newblock A white paper on neural network quantization.
\newblock {\em arXiv preprint arXiv:2106.08295}, 2021.

\bibitem{ramesh2022hierarchical}
Aditya Ramesh, Prafulla Dhariwal, Alex Nichol, Casey Chu, and Mark Chen.
\newblock Hierarchical text-conditional image generation with clip latents.
\newblock {\em arXiv preprint arXiv:2204.06125}, 1(2):3, 2022.

\bibitem{saharia2022photorealistic}
Chitwan Saharia, William Chan, Saurabh Saxena, Lala Li, Jay Whang, Emily~L Denton, Kamyar Ghasemipour, Raphael Gontijo~Lopes, Burcu Karagol~Ayan, Tim Salimans, et~al.
\newblock Photorealistic text-to-image diffusion models with deep language understanding.
\newblock {\em Advances in neural information processing systems}, 35:36479--36494, 2022.

\bibitem{salimans2016improved}
Tim Salimans, Ian Goodfellow, Wojciech Zaremba, Vicki Cheung, Alec Radford, and Xi~Chen.
\newblock Improved techniques for training gans.
\newblock {\em Advances in neural information processing systems}, 29, 2016.

\bibitem{llamagen}
Peize Sun, Yi~Jiang, Shoufa Chen, Shilong Zhang, Bingyue Peng, Ping Luo, and Zehuan Yuan.
\newblock Autoregressive model beats diffusion: Llama for scalable image generation.
\newblock {\em arXiv preprint arXiv:2406.06525}, 2024.

\bibitem{tian2024visual}
Keyu Tian, Yi~Jiang, Zehuan Yuan, Bingyue Peng, and Liwei Wang.
\newblock Visual autoregressive modeling: Scalable image generation via next-scale prediction.
\newblock {\em arXiv preprint arXiv:2404.02905}, 2024.

\bibitem{xiao2023smoothquant}
Guangxuan Xiao, Ji~Lin, Mickael Seznec, Hao Wu, Julien Demouth, and Song Han.
\newblock Smoothquant: Accurate and efficient post-training quantization for large language models.
\newblock In {\em International Conference on Machine Learning}, pages 38087--38099. PMLR, 2023.

\bibitem{attention_sink}
Guangxuan Xiao, Yuandong Tian, Beidi Chen, Song Han, and Mike Lewis.
\newblock Efficient streaming language models with attention sinks.
\newblock {\em arXiv preprint arXiv:2309.17453}, 2023.

\bibitem{xu2024imagereward}
Jiazheng Xu, Xiao Liu, Yuchen Wu, Yuxuan Tong, Qinkai Li, Ming Ding, Jie Tang, and Yuxiao Dong.
\newblock Imagereward: Learning and evaluating human preferences for text-to-image generation.
\newblock {\em Advances in Neural Information Processing Systems}, 36, 2024.

\bibitem{yao2022zeroquant}
Zhewei Yao, Reza Yazdani~Aminabadi, Minjia Zhang, Xiaoxia Wu, Conglong Li, and Yuxiong He.
\newblock Zeroquant: Efficient and affordable post-training quantization for large-scale transformers.
\newblock {\em Advances in Neural Information Processing Systems}, 35:27168--27183, 2022.

\bibitem{yuan2024ditfastattn}
Zhihang Yuan, Pu~Lu, Hanling Zhang, Xuefei Ning, Linfeng Zhang, Tianchen Zhao, Shengen Yan, Guohao Dai, and Yu~Wang.
\newblock Ditfastattn: Attention compression for diffusion transformer models.
\newblock {\em arXiv preprint arXiv:2406.08552}, 2024.

\bibitem{yuan2022ptq4vit}
Zhihang Yuan, Chenhao Xue, Yiqi Chen, Qiang Wu, and Guangyu Sun.
\newblock Ptq4vit: Post-training quantization for vision transformers with twin uniform quantization.
\newblock In {\em European conference on computer vision}, pages 191--207. Springer, 2022.

\bibitem{zhao2024vidit}
Tianchen Zhao, Tongcheng Fang, Enshu Liu, Wan Rui, Widyadewi Soedarmadji, Shiyao Li, Zinan Lin, Guohao Dai, Shengen Yan, Huazhong Yang, et~al.
\newblock Vidit-q: Efficient and accurate quantization of diffusion transformers for image and video generation.
\newblock {\em arXiv preprint arXiv:2406.02540}, 2024.

\bibitem{atom}
Yilong Zhao, Chien-Yu Lin, Kan Zhu, Zihao Ye, Lequn Chen, Size Zheng, Luis Ceze, Arvind Krishnamurthy, Tianqi Chen, and Baris Kasikci.
\newblock Atom: Low-bit quantization for efficient and accurate llm serving.
\newblock {\em Proceedings of Machine Learning and Systems}, 6:196--209, 2024.

\bibitem{luminaX}
Le~Zhuo, Ruoyi Du, Han Xiao, Yangguang Li, Dongyang Liu, Rongjie Huang, Wenze Liu, Lirui Zhao, Fu-Yun Wang, Zhanyu Ma, et~al.
\newblock Lumina-next: Making lumina-t2x stronger and faster with next-dit.
\newblock {\em arXiv preprint arXiv:2406.18583}, 2024.

\end{thebibliography}


\clearpage

\appendix

\section{Appendix / supplemental material}


\textbf{Visualizing the Sensitivity of various kind of Linear Layers to Bit-Width Reduction.}
\begin{figure}[h]
    \centering
    \includegraphics[width=0.5\linewidth]{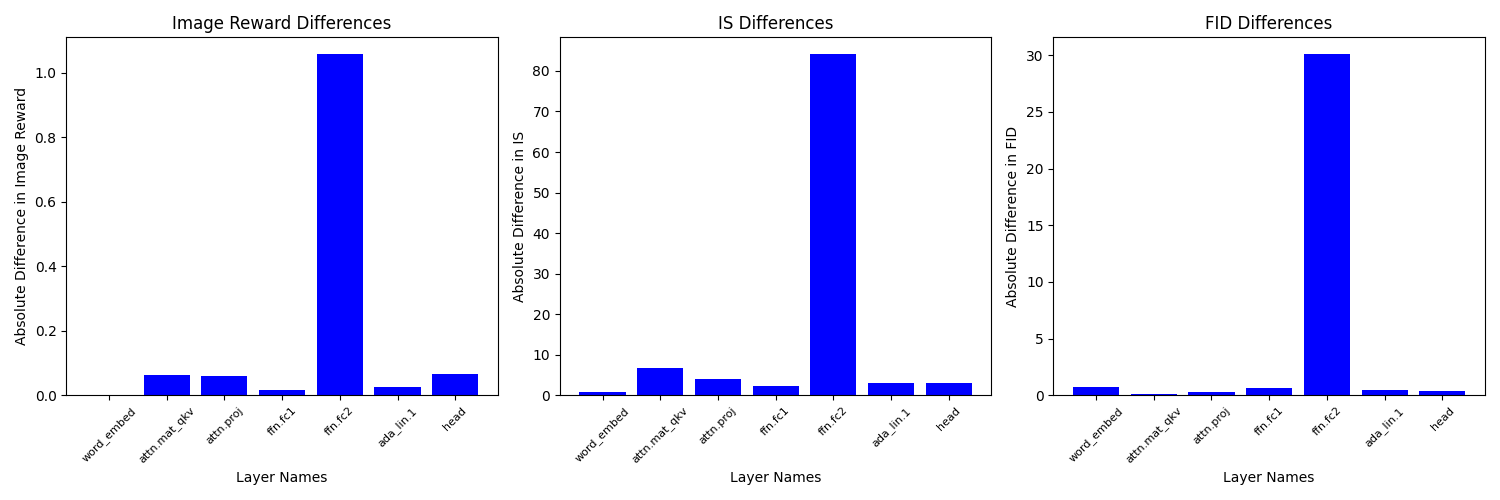}
    \caption{\textbf{Comparison the impact on image quality of all seven types of linear layers:} "\text{word\_\!embed}", "\text{attn.mat\_\!qkv}", "\text{attn.proj}", "\text{ffn.fc1}", "\text{ffn.fc2}", "\text{ada\_\!lin.1}", and "\text{head}". }
    \label{fig:ffn.fc2}
\end{figure}

We observed a particularly noticeable decrease in image quality after quantization for the \textbf{"ffn.fc2"} layer. To address the quantization bottleneck, we have set the bit width of \textbf{ffn.fc2 to FP16} to safeguard sensitive layers.

\textbf{More data for baseline quant:}

\begin{table}[h]
\centering
\caption{\textbf{Performance of image generation on ImageNet under various bitwidths of quantization.}}
\begin{tabular}{@{}cccccccccc@{}}
\toprule[1pt]
\textbf{\begin{tabular}[c]{@{}c@{}}Bit-width\\ (W/A/QKV)\end{tabular}} & \multicolumn{3}{c}{\textbf{FID($\downarrow$)}}   & \multicolumn{3}{c}{\textbf{IS($\uparrow$)}}         & \multicolumn{3}{c}{\textbf{Image reward($\uparrow$)}} \\ \midrule
                                                                       & Original       & Mask           & Cfg            & Original        & Mask            & Cfg             & Original         & Mask             & Cfg             \\ \midrule
16/16/16                                                               & 13.39          & 13.47          & 13.45          & 257.34          & 260.95          & 248.80          & -0.28            & -0.28            & -0.27           \\ \midrule
8/8/16                                                                 & 12.92          & 13.21          & 13.52          & 252.20          & 258.15          & 244.45          & -0.32            & -0.32            & -0.30           \\
8/8/8                                                         & \textbf{12.71} & \textbf{13.02} & \textbf{13.38} & \textbf{249.04} & \textbf{241.02} & \textbf{241.04} & \textbf{-0.33}   & \textbf{-0.37}   & \textbf{-0.29}  \\
4/8/8                                                                  & 54.29          & 56.31          & 52.27          & 40.71           & 33.76           & 33.87           & -1.43            & -1.54            & -1.60           \\
6/6/8                                                                  & 66.53          & 68.89          & 62.09          & 26.08           & 24.57           & 28.13           & -1.68            & -1.73            & -1.72           \\
4/6/8                                                                  & 111.24         & 112.79         & 102.44         & 9.79            & 9.52            & 10.48           & -2.10            & -2.13            & -2.10           \\
4/4/8                                                                  & 133.39         & 134.26         & 139.40         & 6.63            & 6.60            & 5.89            & -2.15            & -2.15            & -2.14           \\ \bottomrule[1pt]
\end{tabular}
\label{tab:3}
\end{table}

We generated 8,000 images on ImageNet to evaluate the quality of our approach. The bitwidth designs for the linear layer portion included W16A16 (the original unquantized model), W8A8, W4A8, W6A6, and W4A6. For the attention computation part, we explored bit-widths of KV8 and KV16. As shown in the table, quantizing the KV section to a bit-width of 8 has minimal impact on image quality. When the quantization precision for the linear layer is set to W8A8KV8, the image quality is comparable to the original floating-point 16-bit (fp16) images. However, W4A8 and W6A6 exhibited significant blurring, and W4A4 resulted in completely illegible images.
Subsequently, we integrated quantization techniques with sparse attention computation to discuss whether the accuracy could still be maintained. As indicated in the table, the image quality degradation after sparse computation and ASC (Attention Sharing across CFG) is minimal, demonstrating that we can significantly reduce computational requirements by approximately 70-90\% while ensuring image quality is preserved.


\textbf{More data for quantization with mixed-precision design to protect sensitive layers:}

Experimental data in \ref{tab:3} reveals that when the weights and activations of linear layers, as well as the attention computation, are set to 8-bit width, the image quality is essentially preserved. However, when the weights and activations are designed with lower bit-widths (e.g., W6A6 or W4A8), the image quality degrades significantly. This is due to the sensitivity of the "ffn.fc2" layer type to quantization, as illustrated in Figure \ref{fig:ffn.fc2}. To address this phenomenon, we set the bit-width of this layer type to fp16, while maintaining the quantized bit-widths for other layers. We can observe that this mixed-precision design significantly improves the performance of W6A6 and W4A8, resulting in noticeably better image quality. For certain metrics (e.g., FID), the W4A8 configuration can even achieve comparable performance to the baseline W8A8 quantization.
\begin{table}[h]
\centering
\caption{Performance of image generation on ImageNet under various settings of quantization. Mixed-precision design significantly improves the performance under low bitwidths quantization.}
\begin{tabular}{@{}cccccccccc@{}}
\toprule[1pt]
\textbf{\begin{tabular}[c]{@{}c@{}}Bit-width\\ (W/A/QKV)\end{tabular}} & \multicolumn{3}{c}{\textbf{FID($\downarrow$)}}   & \multicolumn{3}{c}{\textbf{IS($\uparrow$)}}         & \multicolumn{3}{c}{\textbf{Image reward($\uparrow$)}} \\ \midrule \midrule
                                                                      & Original       & Mask           & Cfg            & Original        & Mask            & Cfg             & Original         & Mask             & Cfg             \\ \midrule
16/16/16                                                              & 13.39          & 13.47          & 13.45          & 257.34          & 260.95          & 248.80          & -0.28            & -0.28            & -0.27           \\ \midrule
8/8/8                                                                 & \textbf{12.71} & \textbf{13.02} & \textbf{13.38} & \textbf{249.04} & \textbf{241.02} & \textbf{241.04} & \textbf{-0.33}   & \textbf{-0.37}   & \textbf{-0.29}  \\
8/8/8+MP                                                             & 13.08          & 13.37          & 13.57          & 253.43          & 251.84          & 244.16          & -0.30            & -0.34            & -0.28           \\
4/8/8                                                                 & 54.29          & 56.31          & 52.27          & 40.71           & 33.76           & 33.87           & -1.43            & -1.54            & -1.60           \\
4/8/8+MP                                                             & \textbf{12.82} & \textbf{13.23} & \textbf{13.34} & \textbf{228.59} & \textbf{229.58} & \textbf{224.74} & \textbf{-0.41}   & \textbf{-0.44}   & \textbf{-0.39}  \\
6/6/8                                                                 & 66.53          & 68.89          & 62.09          & 26.08           & 24.57           & 28.13           & -1.68            & -1.73            & -1.72           \\
6/6/8+MP                                                             & 18.54          & 22.19          & 20.11          & 133.13          & 117.32          & 125.88          & -0.75            & -0.86            & -0.80           \\ \bottomrule[1pt]
\end{tabular}
\label{tab:4}
\end{table}


\textbf{More examples for image generation in different quantization settings.}

Further examples are presented in the following figures, which compare the original image to both the quantized image and the quantized image with the enhanced protection of sensitive layers.

\begin{figure}[h]
    \centering
    \begin{subfigure}[b]{0.22\textwidth}
        \includegraphics[width=\textwidth]{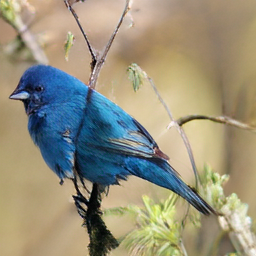}
        \caption{fp16}
    \end{subfigure}%
    \hspace{0pt} 
    \begin{subfigure}[b]{0.22\textwidth}
        \includegraphics[width=\textwidth]{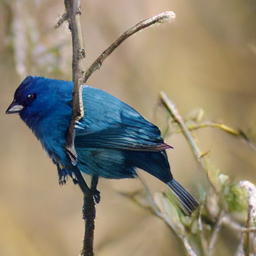}
        \caption{w8a8}
    \end{subfigure}%
    \hspace{0pt} 
    \begin{subfigure}[b]{0.22\textwidth}
        \includegraphics[width=\textwidth]{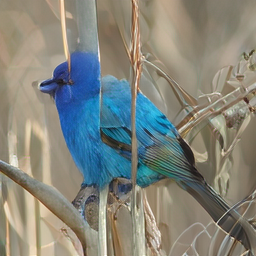}
        \caption{w4a8}
    \end{subfigure}%
    \hspace{0pt} 
    \begin{subfigure}[b]{0.22\textwidth}
        \includegraphics[width=\textwidth]{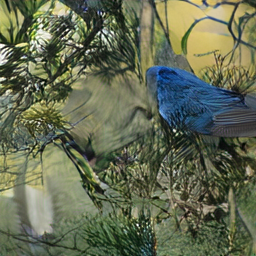}
        \caption{w6a6}
        
    \end{subfigure}

    \vspace{0.2cm} 
    
    \begin{subfigure}[b]{0.22\textwidth}
        \includegraphics[width=\textwidth]{figure/fp16_o_2.png}
        \caption{fp16}
    \end{subfigure}%
    \hspace{0pt} 
    \begin{subfigure}[b]{0.22\textwidth}
        \includegraphics[width=\textwidth]{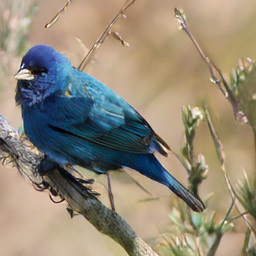}
        \caption{w8a8+MP}
    \end{subfigure}%
    \hspace{0pt} 
    \begin{subfigure}[b]{0.22\textwidth}
        \includegraphics[width=\textwidth]{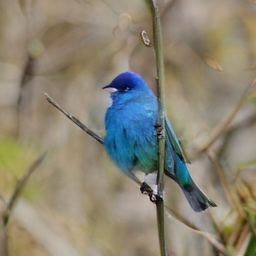}
        \caption{w4a8+MP}
    \end{subfigure}%
    \hspace{0pt} 
    \begin{subfigure}[b]{0.22\textwidth}
        \includegraphics[width=\textwidth]{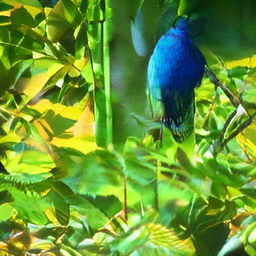}
        \caption{w6a6+MP}
        
    \end{subfigure}
    \caption{More comparison examples.}
    \label{fig:bluebird2}
\end{figure}

\begin{figure}
    \centering
    \begin{subfigure}[b]{0.22\textwidth}
        \includegraphics[width=\textwidth]{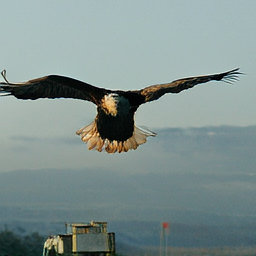}
        \caption{fp16}
    \end{subfigure}%
    \hspace{0pt} 
    \begin{subfigure}[b]{0.22\textwidth}
        \includegraphics[width=\textwidth]{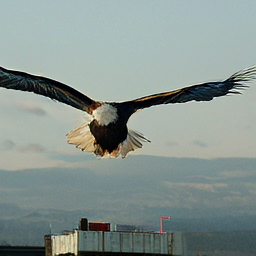}
        \caption{w8a8}
    \end{subfigure}%
    \hspace{0pt} 
    \begin{subfigure}[b]{0.22\textwidth}
        \includegraphics[width=\textwidth]{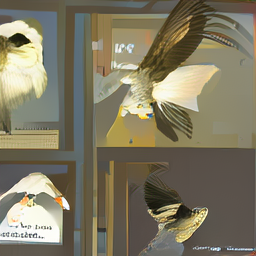}
        \caption{w4a8}
    \end{subfigure}%
    \hspace{0pt} 
    \begin{subfigure}[b]{0.22\textwidth}
        \includegraphics[width=\textwidth]{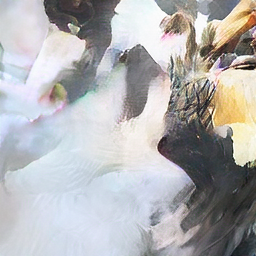}
        \caption{w6a6}
        
    \end{subfigure}

    \vspace{0.2cm} 
    
    \begin{subfigure}[b]{0.22\textwidth}
        \includegraphics[width=\textwidth]{figure/fp16_o_3.png}
        \caption{fp16}
    \end{subfigure}%
    \hspace{0pt} 
    \begin{subfigure}[b]{0.22\textwidth}
        \includegraphics[width=\textwidth]{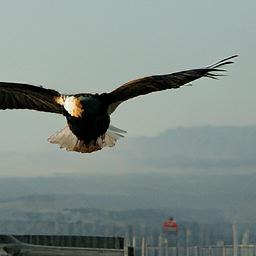}
        \caption{w8a8+MP}
    \end{subfigure}%
    \hspace{0pt} 
    \begin{subfigure}[b]{0.22\textwidth}
        \includegraphics[width=\textwidth]{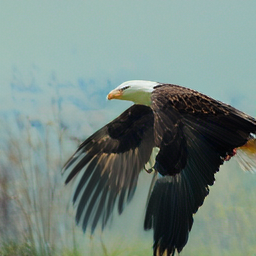}
        \caption{w4a8+MP}
    \end{subfigure}%
    \hspace{0pt} 
    \begin{subfigure}[b]{0.22\textwidth}
        \includegraphics[width=\textwidth]{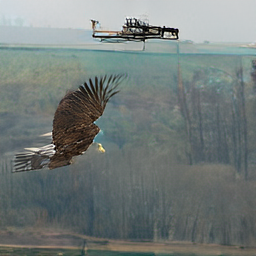}
        \caption{w6a6+MP}
        
    \end{subfigure}
    \caption{More comparison examples.}
    \label{fig:eager3}
\end{figure}

\begin{figure}
    \centering
    \begin{subfigure}[b]{0.22\textwidth}
        \includegraphics[width=\textwidth]{figure/fp16_o_4.png}
        \caption{fp16}
    \end{subfigure}%
    \hspace{0pt} 
    \begin{subfigure}[b]{0.22\textwidth}
        \includegraphics[width=\textwidth]{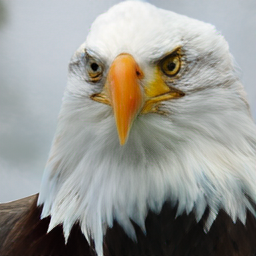}
        \caption{w8a8}
    \end{subfigure}%
    \hspace{0pt} 
    \begin{subfigure}[b]{0.22\textwidth}
        \includegraphics[width=\textwidth]{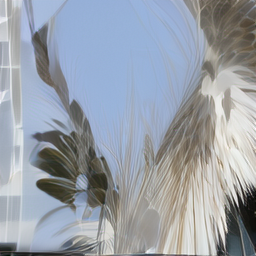}
        \caption{w4a8}
    \end{subfigure}%
    \hspace{0pt} 
    \begin{subfigure}[b]{0.22\textwidth}
        \includegraphics[width=\textwidth]{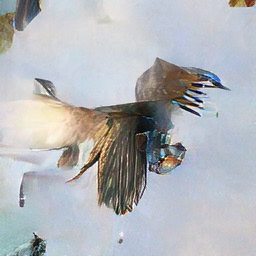}
        \caption{w6a6}
        
    \end{subfigure}

    \vspace{0.2cm} 
    
    \begin{subfigure}[b]{0.22\textwidth}
        \includegraphics[width=\textwidth]{figure/fp16_o_4.png}
        \caption{fp16}
    \end{subfigure}%
    \hspace{0pt} 
    \begin{subfigure}[b]{0.22\textwidth}
        \includegraphics[width=\textwidth]{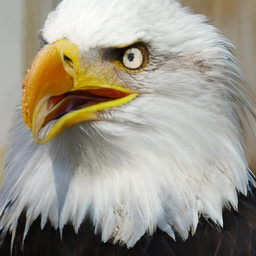}
        \caption{w8a8+MP}
    \end{subfigure}%
    \hspace{0pt} 
    \begin{subfigure}[b]{0.22\textwidth}
        \includegraphics[width=\textwidth]{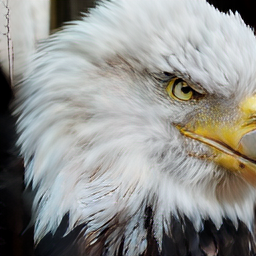}
        \caption{w4a8+MP}
    \end{subfigure}%
    \hspace{0pt} 
    \begin{subfigure}[b]{0.22\textwidth}
        \includegraphics[width=\textwidth]{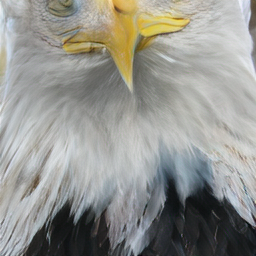}
        \caption{w6a6+MP}
        
    \end{subfigure}
    \caption{More comparison examples.}
    \label{fig:eager4}
\end{figure}


\end{document}